%% file: main.tex
\documentclass[lettersize,journal]{IEEEtran}
\usepackage{amsmath,amsfonts}
\usepackage{algorithmic}
\usepackage{algorithm}
\usepackage{array}
\usepackage[caption=false,font=normalsize,labelfont=sf,textfont=sf]{subfig}
\usepackage{textcomp}
\usepackage{stfloats}
\usepackage{url}
\usepackage{verbatim}
\usepackage{graphicx}
\usepackage{cite}
\usepackage{bm}
\usepackage{orcidlink}
\usepackage{xcolor}
\usepackage{multirow}
\usepackage{makecell}

\IEEEaftertitletext{\vspace{-1.8\baselineskip}}

\begin{document}
\title{SwiftQK: Fast and Communication-Efficient Tensor Parallelism for Query-Key Normalization}
\author{Gyudong Kim\orcidlink{0009-0006-3542-9722}, Wonjun Han\orcidlink{0009-0008-5364-3452}, and Young Geun Kim\orcidlink{0000-0003-4713-819X}
\thanks{This work was supported in part by the National Research Foundation of Korea (NRF) grant funded by MSIT (RS-2025-24534857, RS-2026-25522655, RS-2025-25434746, and RS-2026-25486583), in part by IITP-ITRC (Information Technology Research Center) grant funded by MSIT under Grant IITP-2026-RS-2023-00260091, and in part by IITP-ICT Creative Consilience Program grant funded by MSIT under Grant IITP-2026-RS-2020-II201819. \textit{(Gyudong Kim and Wonjun Han are co-first authors.) (Corresponding author: Young Geun Kim.)}}

\thanks{Department of Computer Science and Engineering, Korea University, Seoul 02855, Republic of Korea (e-mail:
gyudong\_kim@korea.ac.kr; siftheads@korea.ac.kr; younggeun\_kim@korea.ac.kr).}}

\markboth{IEEE Computer Architecture Letters}%
{Shell \MakeLowercase{\textit{et al.}}: A Sample Article Using IEEEtran.cls for IEEE Journals}

\maketitle

\input{contents-R2/00Abstract}


\input{contents-R2/01Introduction}

\input{contents-R2/02BackgroundMotivation}

\input{contents-R2/05Design}

\input{contents-R2/06Evaluation}
\input{contents-R2/07Conclusion}

\vspace{-5pt}
\bibliographystyle{IEEEtran}
\bibliography{references}

\end{document}

%% file: contents-R2/00Abstract.tex
\begin{abstract}
Query-Key Normalization (QK-Norm) improves the training stability and quality of modern Large Language Models (LLMs).
However, under Tensor Parallelism (TP), layerwise QK-Norm introduces additional cross-GPU communication because the normalization factor depends on the full hidden vector.
We present SwiftQK, a multi-GPU RMSNorm kernel that exchanges only scalar normalization statistics and overlaps the remaining Peer-to-Peer reduction with independent element-wise computation in a deadlock-safe persistent kernel.
Evaluations on recent LLMs show that SwiftQK reduces QK-Norm latency by 81.4--93.9\% relative to the standard TP QK-Norm using full-vector All-Gather.
In end-to-end serving, SwiftQK reduces TPOT on average by 29.5\% over the All-Gather-based baseline and by 14.3\% over an optimized scalar-aggregation implementation.
\end{abstract}
\makeatletter
\def\subsection{\@startsection{subsection}{2}{\z@}{1.5ex plus 0.5ex minus 0.5ex}%
{0.7ex plus .5ex minus 0ex}{\normalfont\normalsize\itshape}}
\makeatother

%% file: contents-R2/01Introduction.tex
\section{Introduction}
\IEEEPARstart{T}{he} rapid growth of Large Language Models (LLMs) has made distributed serving essential for addressing limited GPU memory capacity and latency constraints~\cite{shoeybi2019megatron}.
A widely adopted strategy for distributed serving is Tensor Parallelism (TP), which shards intra-layer computation across multiple GPUs.
By parallelizing each layer, TP reduces per-request latency.
As a result, modern inference engines such as vLLM~\cite{kwon2023efficient} commonly adopt TP.

Recent LLMs increasingly adopt Query-Key Normalization (QK-Norm) to stabilize attention.
In its layerwise formulation, each normalization factor depends on the full projected Query (Q) or Key (K) vector.
However, this vector is partitioned across GPUs under TP, requiring cross-GPU synchronization that introduces substantial communication overhead.

To mitigate the communication overhead of TP, existing kernel-fusion techniques overlap communication with computation~\cite{chang2024flux,hong2025flashoverlap}.
These techniques are effective when communication can be hidden behind sufficient independent computation.
However, QK-Norm has lightweight computation and large cross-GPU synchronization cost, making overlap alone insufficient.
This motivates combining communication-volume reduction with overlapping for layerwise QK-Norm under TP.

In this letter, we propose \textbf{SwiftQK}, a communication-efficient multi-GPU RMS-Norm kernel for QK-Normalization under TP.
SwiftQK addresses a new operator-level bottleneck in TP by reformulating QK-Norm as an operator-aware communication-minimal primitive, rather than only as a communication-overlap problem. 
It replaces full-vector Q/K activation exchange with scalar partial-sum aggregation and overlaps Peer-to-Peer (P2P) scalar reduction with independent element-wise multiplication over the same local Q/K shard inside a deadlock-safe persistent kernel.

Evaluations on recent open language models show that SwiftQK reduces TPOT by 29.5\% and increases saturated request throughput by 25.4\% on average over the All-Gather-based QK-Norm used by the evaluated models.
Notably, even against an optimized scalar-aggregation implementation, SwiftQK reduces TPOT by 14.3\% and increases saturated request throughput by 8.8\% on average.

%% file: contents-R2/02BackgroundMotivation.tex
\section{Background and Motivation}
\begin{figure*}[t]
    \centering
    \includegraphics[width=1.0\linewidth]{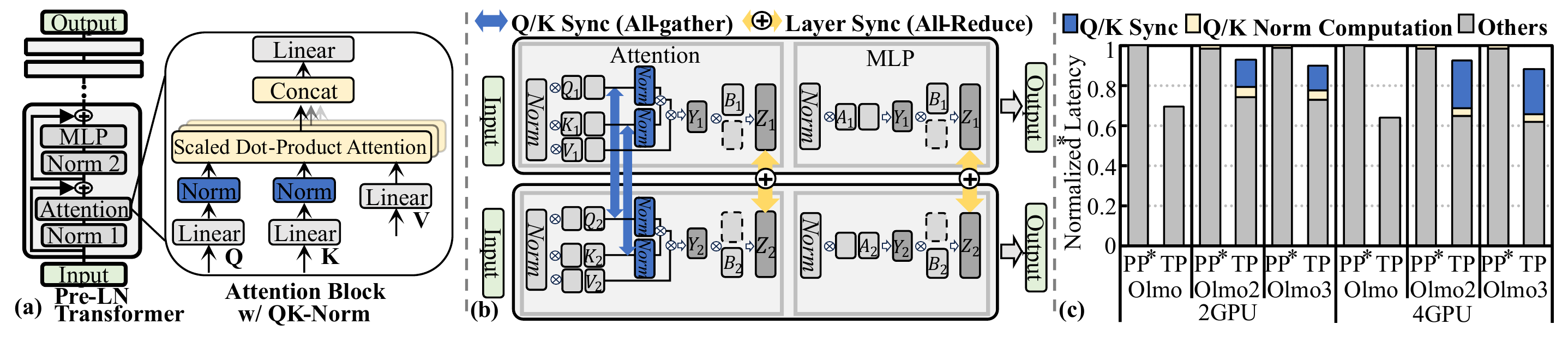}
    \vspace{-20pt}
    \caption{(a) Transformer block w/QK-Norm (b) TP w/layerwise QK-Norm, requiring All-Gather Q/K sync (c) PP--TP normalized total prompt-processing latency comparison w/ and w/o layerwise QK-Norm; for each model and GPU count, latency is normalized to the corresponding PP latency.}
    \label{fig:back_all}
    \vspace{-16pt}
\end{figure*}
\subsection{Layerwise QK-Normalization}
Earlier LLMs followed a standard Pre-LN transformer architecture.
In this design, linear projections produce the Query (Q), Key (K), and Value (V) tensors, which are then directly passed to Scaled Dot-Product Attention.

As model sizes and input lengths increase, stabilizing attention logits becomes more important.
To address this issue, ViT-22B~\cite{dehghani2023scaling} adopts Query-Key Normalization (QK-Norm), which normalizes the projected Q and K tensors before attention.

QK-Norm can be applied at different granularities.
Layerwise QK-Norm computes one normalization factor over the entire projected Q or K dimension, whereas headwise QK-Norm computes an independent factor for each attention head.
Recent open language model series~\cite{Olmo2,Olmoe,Olmo3} adopt the layerwise formulation by applying RMSNorm~\cite{zhang2019root} before Scaled Dot-Product Attention (Fig.~\ref{fig:back_all}(a)).

Let $x \in \mathbb{R}^H$ denote a full projected Q or K vector, where $H$ is its dimension.
RMS-Norm is computed as:
\begin{equation}
\label{eq:rmsnorm} 
\text{RMSNorm}(x) = \frac{x}{\sqrt{\frac{1}{H} \sum_{j=1}^{H} x_j^2 + \epsilon}} \odot \gamma 
\end{equation}
where $\epsilon$ is a small constant for numerical stability and $\gamma$ is a learnable scaling parameter.
Because the normalization factor depends on all $H$ elements, partitioning this dimension under TP requires cross-rank aggregation.

\subsection{Model Parallelism}
Serving large language models often requires distributing computation and memory across multiple GPUs.
Inference systems therefore rely on model parallelism, which partitions a model across devices to reduce per-device resource demands.
The two common forms are Pipeline Parallelism (PP)~\cite{huang2019gpipe} and Tensor Parallelism (TP)~\cite{shoeybi2019megatron}.
PP partitions the model by layers, whereas TP partitions each layer across GPUs.

\subsubsection{Pipeline Parallelism}
In PP, GPUs execute different groups of layers as sequential pipeline stages.
This reduces the per-device model footprint and limits communication to stage boundaries.
However, PP can introduce pipeline bubbles and inter-stage dependencies, which increase serving latency.

\subsubsection{Tensor Parallelism}
For latency-sensitive LLM serving, modern inference engines~\cite{kwon2023efficient} often rely on TP when fast inter-GPU interconnects are available.
In TP, GPUs jointly execute each layer by sharding intra-layer operations such as linear projections and attention.
As shown in Fig.~\ref{fig:back_all}(b), the first linear projections in Attention and MLP layers are partitioned along the output dimension, while the following projections are partitioned along the input dimension.
This layout allows each GPU to directly consume its local activation shard in the next matrix multiplication, requiring synchronization only when the final projection output is combined, typically via All-Reduce.
By avoiding pipeline bubbles and unnecessary communication between consecutive matrix multiplications, TP is effective for reducing per-request latency.

\subsection{Characterization of Layerwise QK-Norm Overhead}
Recent LLMs increasingly adopt QK-Norm to stabilize attention.
However, layerwise QK-Norm breaks the communication-efficient execution pattern of TP by inserting a normalization step between the Q/K projections and attention computation.
Since TP partitions the projected Q/K hidden dimension across GPUs, each GPU only holds a shard of the full projected Q or K vector.
RMS-Norm requires a reduction over the full hidden vector, so each GPU cannot compute the correct normalization factor from its local shard alone.
This introduces synchronization of Q/K vectors before attention can proceed, typically via All-Gather in TP (Fig.~\ref{fig:back_all}(b)).

Fig.~\ref{fig:back_all}(c) reports normalized total prompt-processing time for ShareGPT request set~\cite{sharegpt_vicuna_unfiltered} on OLMo~\cite{Olmo}, OLMo2~\cite{Olmo2}, and OLMo3~\cite{Olmo3}. The workload uses vLLM's unified serving mode with continuous batching and is evaluated on two and four A100 GPUs under PP and TP.
The earlier model OLMo does not use QK-Norm, whereas OLMo2 and OLMo3 apply it to the projected Q and K tensors.
The breakdown separates Q/K synchronization, Q/K normalization computation, and the remaining execution time.
Across all settings, TP achieves lower latency than PP, and the benefit generally increases with more GPUs.
However, this benefit is much smaller for OLMo2 and OLMo3 than for OLMo: on four GPUs, TP reduces latency by 36.2\% for OLMo, but only by 7.5\% and 12.0\% for OLMo2 and OLMo3, respectively.
This gap shows that QK-Norm significantly weakens the latency advantage of TP.

Under PP, QK-Norm overhead (Q/K synchronization + Q/K normalization computation) contributes only a small fraction of the total execution time, accounting for 1.4\% and 1.3\% on average for OLMo2 and OLMo3, respectively.
In contrast, under TP, QK-Norm overhead becomes a major bottleneck, and its cost grows with the TP degree.
For OLMo2 and OLMo3, its overhead increases from 20.0\% and 19.0\% on two GPUs to 30.1\% and 29.7\% on four GPUs, respectively.
This increase is dominated by Q/K synchronization, which accounts for 73.6\% and 72.8\% of QK-Norm overhead on two GPUs and rises to 85.6\% and 85.5\% on four GPUs.
This trend occurs because increasing the TP degree reduces per-GPU normalization computation, but increases exposed synchronization cost.
As a result, layerwise QK-Norm overhead becomes a communication-dominated bottleneck that limits TP efficiency.

\subsection{Kernel Fusion for Communication Overlap}
To reduce exposed communication overhead of TP, existing kernel fusion methods such as FLUX~\cite{chang2024flux} and FlashOverlap~\cite{hong2025flashoverlap} overlap communication with computation.
These methods mainly target GEMM-related communication patterns, such as GEMM followed by All-Reduce, where matrix multiplication provides enough computation to hide communication latency.

However, overlap is effective only when there is sufficient computation to cover the communication cost.
This assumption does not hold for QK-Normalization under TP (see Fig.~\ref{fig:eval_bench} for comparison).
RMS-Norm performs only lightweight reduction and element-wise operations, while the required cross-GPU communication cost is large.
Therefore, simply overlapping communication with computation is insufficient to remove the QK-Norm bottleneck under TP.

%% file: contents-R2/05Design.tex
\section{Design and Implementation of SwiftQK} \label{sec:proposed-method}
To mitigate the layerwise QK-Norm bottleneck under TP, we propose \textbf{SwiftQK}, a communication-efficient fused multi-GPU RMS-Norm kernel.
SwiftQK performs the entire QK-Norm procedure within a single fused kernel, reducing the kernel launch overhead caused by separate communication and normalization kernels.
For each token, SwiftQK decomposes RMS-Norm into three phases: local squared-sum computation, communication-computation overlap between peer-to-peer reduction and element-wise multiplication, and final normalization with the global RMS factor.
SwiftQK executes these phases in a persistent fused kernel, where resident CUDA blocks repeatedly process tokens in a block-level token pipeline.
The algorithm detail is explained in Algorithm~\ref{alg:fused_rms_norm}.

\subsection{Phase A: Local Aggregation for Communication Reduction}

The first phase of SwiftQK reduces QK-Norm communication to the minimum statistic needed for RMS-Norm.
In standard TP execution, each GPU exchanges full Q and K activation shards through All-Gather before computing the normalization statistics.
However, the RMS factor (denominator of Equation~\ref{eq:rmsnorm}) does not require the full hidden vector.
It only requires the squared sum of the hidden vector, $\sum_{j=1}^{H} x_j^2$.

SwiftQK computes a local squared sum on each GPU over its hidden-dimension shard and aggregates these partial sums across GPUs using FP32 accumulation.
This recovers the global squared sum required for the RMS factor, replacing the $O(H)$ activation exchange with $O(1)$ scalar partial-sum aggregation while preserving the same global normalization semantics.

\subsection{Phase B: Comm--Comp Overlap for Latency Hiding}

Although Phase A reduces the communication volume, the P2P reduction still requires synchronization across GPUs.
Each GPU can compute the global squared sum only after the scalar partial sums from all peer GPUs become visible through the IPC buffers.
This waiting time remains as synchronization latency even after the communication volume is reduced.

SwiftQK hides this remaining latency by overlapping the P2P reduction with RMS-Norm weight multiplication, which is independent of the reduction result.
Specifically, Warp 0 is dedicated to the communication path and performs the P2P reduction over the IPC buffers to obtain the global squared sum.
Meanwhile, the remaining warps perform element-wise multiplication with the RMS-Norm weights.
Because this multiplication does not depend on the global squared sum, it can safely proceed while the P2P reduction is in progress.

\subsection{Phase C: Final Normalization with the Global RMS Factor}

After the P2P reduction and the element-wise multiplication complete, SwiftQK has all inputs needed for the RMS scaling: the global squared sum and the weight-multiplied local shard.
It then applies the global RMS scaling factor to each local element and writes only the final normalized shard to HBM.

\subsection{Deadlock-Safe Persistent Execution}

All peer blocks participating in the same synchronization step must be concurrently resident during in-kernel P2P synchronization.
If one block waits for a peer block that is not yet resident, the kernel can enter cyclic waiting, i.e., deadlock.

To prevent deadlock, SwiftQK uses a persistent execution model with a bounded launch grid.
It launches only as many blocks as can reside concurrently on the Streaming Multiprocessors (SMs), rather than launching one block per token.
Let $B_{\mathrm{res}}$ denote the maximum number of resident blocks per SM under the SwiftQK kernel configuration, and let $N_{\mathrm{SM}}$ denote the number of SMs.
SwiftQK launches at most $B_{\mathrm{res}} \times N_{\mathrm{SM}}$ blocks, where $B_{\mathrm{res}}$ is determined by the per-block resource usage, including registers, shared memory, and thread count.

When the number of tokens exceeds the number of resident blocks, each block processes multiple tokens in a loop, forming a block-level token pipeline.
This preserves deadlock safety while maintaining throughput without launching separate communication and normalization kernels.

\renewcommand{\algorithmicrequire}{\textbf{Input:}}
\renewcommand{\algorithmicensure}{\textbf{Output:}}
\setlength{\textfloatsep}{0pt plus 1pt minus 1pt}
\setlength{\floatsep}{0pt plus 1pt minus 1pt}

\begin{algorithm}[t]
\caption{Fused Multi-GPU Persistent RMS-Norm Kernel}
\label{alg:fused_rms_norm}
\small
\begin{algorithmic}
\REQUIRE Local Input $X_{local}$, Weights $W_{local}$, Total Hidden $H_{total}$
\REQUIRE Num GPUs $N$, Current Rank $r$, IPC Buffers $B_{IPC}$
\ENSURE Normalized local shard $Y_{local}$

\FOR{each token $t$ in assigned tokens (stride loop)}
    \STATE \colorbox{gray!20}{\strut\textbf{Phase A: Local Sum of Squares}}
    \STATE $S_{local} \leftarrow \mathrm{BlockReduce\_Sum}(X_{local}[t]^2)$
    \IF{$\mathrm{thread\_id} = 0$}
        \STATE $B_{IPC}[r].\mathrm{sum_{sq}}[t] \leftarrow S_{local}$
    \ENDIF
    \STATE \texttt{\_\_syncthreads()} \hfill // Barrier for Phase A

    \STATE \colorbox{gray!20}{\strut\textbf{Phase B: Comm-Comp Overlap}}
    \IF{$\mathrm{warp\_id} = 0$}
        \STATE \textit{// Communication Path: Push-wait sync}
        \STATE RemoteWrite(\textit{flag}) to peer GPUs
        \STATE SpinWait(\textit{local\_flag})
        \STATE $S_{global} \leftarrow \sum_{k=0}^{N-1} B_{IPC}[k].\mathrm{sum_{sq}}[t]$ \hfill // P2P reduction
        \STATE $RMS^{-1} \leftarrow 1 / \sqrt{S_{global} / H_{total} + \epsilon}$
    \ELSE
        \STATE \textit{// Computation Path: Element-wise Product}
        \FOR{each $j$ in assigned hidden dimensions}
            \STATE $X_{local}[t][j] \leftarrow X_{local}[t][j] \times W_{local}[j]$
        \ENDFOR
    \ENDIF
    \STATE \texttt{\_\_syncthreads()} \hfill // Barrier for fused ops

    \STATE \colorbox{gray!20}{\strut\textbf{Phase C: Final Normalization}}
    \FOR{each $j$ in assigned hidden dimensions}
        \STATE $Y_{local}[t][j] \leftarrow X_{local}[t][j] \times RMS^{-1}$
    \ENDFOR
\ENDFOR
\end{algorithmic}
\end{algorithm}

%% file: contents-R2/06Evaluation.tex
\section{Evaluation}
\subsection{Experimental Setup}
\noindent \textbf{Models and Workloads:}
We evaluate SwiftQK on three recent LLMs with layerwise QK-Norm from the open language model series: OLMoE (7B, MoE)~\cite{Olmoe}, OLMo 2 (13B, dense), and OLMo 3 (32B, dense).
This selection varies model scale and architecture while keeping QK-Norm common across models.
We use ShareGPT dataset and conduct model-specific RPS sweeps below saturation: 46.5--48 RPS for OLMoE, 23.5--25 RPS for OLMo 2, and 12.5--14 RPS for OLMo 3.
For the saturated-throughput evaluation, we submit the complete request set at once to operate the server under saturating load.

\noindent \textbf{Execution Environment:}
We conduct micro-architectural profiling on two NVLink-connected NVIDIA RTX 3090 GPUs and end-to-end serving evaluation on 4- and 8-GPU NVLink-connected NVIDIA A100 servers.
All kernels are integrated into the vLLM serving engine.

\noindent \textbf{Baselines and Metrics:}
We compare SwiftQK with All-Gather-based QK-Norm, \textit{Comm-Overlap}, and MiniMax(eager), and MiniMax(fusion).
\textit{Comm-Overlap} overlaps All-Gather communication with RMSNorm computation without reducing the communication payload. 
MiniMax(eager) performs scalar-statistic aggregation without full-vector Q/K All-Gather, while MiniMax(fusion) further fuses scalar-statistic aggregation and RMSNorm computation into an optimized kernel~\cite{vllm_minimax_rms_norm_tp}.
In both MiniMax baselines, cross-rank scalar synchronization and RMSNorm computation are executed sequentially, without hiding the synchronization latency behind computation.
We report TPOT and saturated request throughput, defined as the average interval between output tokens and completed requests per second, respectively.

\begin{figure*}[t]
    \centering
    \includegraphics[width=1.0\linewidth]{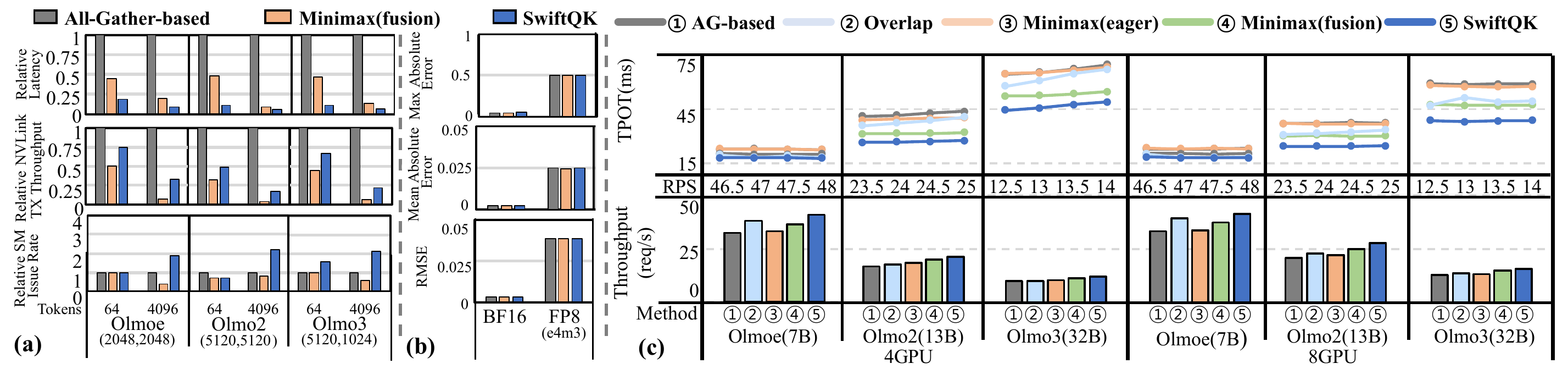}
    \vspace{-20pt}
    \caption{(a) Micro-architectural profiling; values in ( , ) indicate Q/K sizes. (b) Numerical precision comparison. (c) Serving performance comparison.}
    \label{fig:eval_bench}
    \vspace{-16pt}
\end{figure*}

\subsection{Micro-architectural Profiling}

Fig.~\ref{fig:eval_bench}(a) compares All-Gather-based QK-Norm, MiniMax(fusion), and SwiftQK at 64 and 4096 input tokens.
The reported latency includes both communication and normalization computation, and all metrics are normalized to All-Gather-based QK-Norm.
Across all evaluated models and token counts, SwiftQK reduces QK-Norm latency by 81.4--93.9\% compared with All-Gather-based QK-Norm and by 29.4--77.0\% compared with MiniMax(fusion).
These results show that SwiftQK provides additional latency reduction beyond scalar-statistic aggregation.

Both MiniMax(fusion) and SwiftQK show lower measured NVLink TX throughput than All-Gather-based QK-Norm, reflecting their compact scalar communication payloads.
At 4096 tokens, SwiftQK achieves a 2.8--4.6$\times$ higher SM issue rate than MiniMax(fusion). 
This result is consistent with SwiftQK’s in-kernel communication-computation overlap, where Warp 0 performs scalar P2P reduction while the remaining warps execute independent weight multiplication.

\subsection{Numerical Precision Comparison}
Fig.~\ref{fig:eval_bench}(b) compares the numerical precision of the All-Gather-based QK-Norm, MiniMax(fusion), and SwiftQK against a high-precision gold output computed by full All-Gather followed by FP64 RMSNorm.
SwiftQK accumulates local squared sums and cross-GPU scalar reductions in FP32, even for BF16 and FP8-E4M3 activations. 
SwiftQK closely matches the reference path in maximum absolute, mean absolute, and RMS errors: \(5.0\mathrm{e}-2\), \(2.1\mathrm{e}-3\), and \(3.4\mathrm{e}-3\) for BF16, and \(5.0\mathrm{e}-1\), \(2.5\mathrm{e}-2\), and \(3.9\mathrm{e}-2\) for FP8-E4M3, respectively. 
These results indicate that SwiftQK does not introduce additional numerical error beyond the target low-precision format.

\subsection{End-to-End Serving Performance}
Fig.~\ref{fig:eval_bench}(c) compares TPOT across model-specific RPS sweeps and saturated request throughput. 
Compared with the standard All-Gather-based baseline, SwiftQK reduces TPOT by 29.5\% and increases saturated request throughput by 25.4\% on average.
Compared with Comm-Overlap and MiniMax(eager), SwiftQK reduces TPOT by 17.9\% and 28.5\%, respectively, with corresponding throughput gains of 14.6\% and 20.1\%.
Notably, even against an optimized scalar-aggregation implementation, SwiftQK reduces TPOT by 14.3\% and increases saturated request throughput by 8.8\% on average.
These results show that SwiftQK's end-to-end benefit is not only from scalar-statistic aggregation, but also from combining reduced communication volume with fused persistent execution and in-kernel communication-computation overlap.

%% file: contents-R2/07Conclusion.tex
\section{Conclusion and future work}
SwiftQK is a communication-efficient multi-GPU RMS-Norm kernel for layerwise QK-Normalization under TP.
It replaces full-vector activation exchange with scalar partial-sum aggregation and overlaps Peer-to-Peer reduction with independent element-wise computation, reducing QK-Norm communication without changing normalization semantics.
On recent OLMo models, SwiftQK reduces QK-Norm latency by 81.4--93.9\% relative to the All-Gather-based QK-Norm.
In end-to-end serving, SwiftQK reduces TPOT on average by 29.5\% over All-Gather-based QK-Norm and by 14.3\% over an optimized scalar-aggregation implementation.

There are various ways to place and define normalization within Transformer blocks, and recent work continues to explore this design space~\cite{chen2026simplegpt}. 
When normalization spans TP-partitioned activations and requires cross-GPU synchronization, SwiftQK’s design principles may also apply.
Future work will explore how to adapt SwiftQK to these normalization variants.

%% file: references.bib
@inproceedings{kwon2023efficient,
  title={Efficient memory management for large language model serving with pagedattention},
  author={Kwon, Woosuk and others},
  booktitle={Proceedings of the 29th symposium on operating systems principles},
  pages={611--626},
  year={2023}
}

@inproceedings{dehghani2023scaling,
  title={Scaling vision transformers to 22 billion parameters},
  author={Dehghani, Mostafa and others},
  booktitle={International conference on machine learning},
  pages={7480--7512},
  year={2023},
  organization={PMLR}
}

@article{chang2024flux,
  title={Flux: Fast software-based communication overlap on gpus through kernel fusion},
  author={Chang, Li-Wen and others},
  journal={arXiv preprint arXiv:2406.06858},
  year={2024}
}

@article{hong2025flashoverlap,
  title={Flashoverlap: A lightweight design for efficiently overlapping communication and computation},
  author={Hong, Ke and others},
  journal={arXiv preprint arXiv},
  volume={2504},
  year={2025}
}

@misc{sharegpt_vicuna_unfiltered,
  author       = {{anon8231489123}},
  title        = {{ShareGPT Dataset}},
  year         = {2023},
  howpublished = {\url{https://huggingface.co/datasets/anon8231489123/ShareGPT_Vicuna_unfiltered}},
}

@inproceedings{Olmo,
  title={OLMo: Accelerating the science of language models},
  author={Groeneveld, Dirk and others},
  booktitle={Proceedings of the 62nd Annual Meeting of the Association for Computational Linguistics},
  pages={15789--15809},
  year={2024}
}

@article{Olmo2,
  title={2 OLMo 2 Furious},
  author={OLMo, Team},
  journal={arXiv preprint arXiv:2501.00656},
  year={2024}
}

@article{Olmo3,
  title={Olmo 3},
  author={Olmo, Team},
  journal={arXiv preprint arXiv:2512.13961},
  year={2025}
}

@inproceedings{Olmoe,
  title={Olmoe: Open mixture-of-experts language models},
  author={Muennighoff, Niklas and others},
  booktitle={International Conference on Learning Representations},
  volume={2025},
  pages={62061--62121},
  year={2025}
}

@article{shoeybi2019megatron,
  title={Megatron-lm: Training multi-billion parameter language models using model parallelism},
  author={Shoeybi, Mohammad and others},
  journal={arXiv preprint arXiv:1909.08053},
  year={2019}
}

@article{huang2019gpipe,
  title={Gpipe: Efficient training of giant neural networks using pipeline parallelism},
  author={Huang, Yanping and others},
  journal={Advances in neural information processing systems},
  volume={32},
  year={2019}
}

@article{zhang2019root,
  title={Root mean square layer normalization},
  author={Zhang, Biao and Sennrich, Rico},
  journal={Advances in neural information processing systems},
  volume={32},
  year={2019}
}

@misc{vllm_minimax_rms_norm_tp,
  author       = {{vLLM Project}},
  title        = {{MiniMax TP RMSNorm}},
  year         = {2026},
  howpublished = {\url{https://github.com/vllm-project/vllm}},
  note         = {rms\_norm\_tp.py, commit 99a8561}
}

@article{chen2026simplegpt,
  title={SimpleGPT: Improving GPT via A Simple Normalization Strategy},
  author={Chen, Marco and others},
  journal={arXiv preprint arXiv:2602.01212},
  year={2026}
}
